\documentclass[runningheads]{llncs}
\usepackage[T1]{fontenc}
\usepackage{graphicx}

\usepackage{amsfonts}
\usepackage{amsmath}
\usepackage{amssymb}
\usepackage{bm, bbm}
\usepackage[capitalize]{cleveref}
\usepackage[inline]{enumitem}
\usepackage{mathtools}
\usepackage{nicefrac}
\usepackage{tikz}
\usepackage{xspace}

\newcommand{\QUBO}{\textsc{Qubo}\xspace}
\newcommand{\range}[1]{\left[ #1 \right]}

\newcommand{\bQ}{\bm{Q}}

\newcommand{\bw}{\bm{w}}
\newcommand{\bX}{\bm{X}}
\newcommand{\bx}{\bm{x}}
\newcommand{\by}{\bm{y}}
\newcommand{\bz}{\bm{z}}
\newcommand{\BB}{\mathbb{B}}
\newcommand{\BN}{\mathbb{N}}
\newcommand{\BR}{\mathbb{R}}
\newcommand{\BS}{\mathbb{S}}
\newcommand{\cD}{\mathcal{D}}

\newcommand{\T}{\intercal}

\DeclarePairedDelimiter\norm\lVert\rVert
\DeclarePairedDelimiter\set\lbrace\rbrace

\begin{document}
\title{On the Impact of Weight Discretization in QUBO-Based SVM Training}
\author{Sascha M\"ucke\inst{1}\orcidID{0000-0001-8332-61693}}
\authorrunning{S. M\"ucke}
\institute{Lamarr Institute, Dortmund, Germany\\\email{sascha.muecke@tu-dortmund.de}}
\maketitle              
\begin{abstract}
Training Support Vector Machines (SVMs) can be formulated as a QUBO problem, enabling the use of quantum annealing for model optimization.
In this work, we study how the number of qubits---linked to the discretization level of dual weights---affects predictive performance across datasets.
We compare QUBO-based SVM training to the classical LIBSVM solver and find that even low-precision QUBO encodings (e.g., 1 bit per parameter) yield competitive, and sometimes superior, accuracy.
While increased bit-depth enables larger regularization parameters, it does not always improve classification.
Our findings suggest that selecting the right support vectors may matter more than their precise weighting.
Although current hardware limits the size of solvable QUBOs, our results highlight the potential of quantum annealing for efficient SVM training as quantum devices scale.
\end{abstract}
\section{Introduction}

Support Vector Machines (SVMs) are a well-established class of supervised learning models used for classification and regression tasks~\cite{cortes.vapnik.1995a,vapnik.2000a}.
Known for their solid theoretical foundations and effectiveness in high-dimensional spaces, SVMs have found applications in bioinformatics~\cite{byvatov.schneider.2003a}, image recognition~\cite{chandra.bedi.2021a}, and text categorization~\cite{joachims.1998a}, to name a few.

Quantum optimization is an emerging research area with the potential to surpass classical optimization approaches in certain problem domains.
Quantum Annealing (QA) is a metaheuristic that exploits quantum-mechanical phenomena such as superposition and quantum tunneling to search for low-energy configurations of a system~\cite{kadowaki.nishimori.1998a}.
The algorithm begins with a quantum system in the ground state of a simple initial Hamiltonian.
Over time, this Hamiltonian is slowly evolved into a problem-specific Hamiltonian that encodes the cost function of interest, typically formulated as a Quadratic Unconstrained Binary Optimization (\QUBO) problem.
Due to quantum tunneling, QA can escape local minima that would trap classical methods like simulated annealing, potentially leading to improved solutions for certain classes of hard combinatorial problems.
\QUBO is an NP-hard formulation that has been widely used to encode real-world optimization tasks for quantum annealers~\cite{lucas.2014a,muecke.etal.2019a,date.etal.2020a,muecke.etal.2023a,muecke.gerlach.2023a,muecke.2024a}.
However, the practical utility of QA is currently constrained by hardware limitations, including restricted qubit counts, limited connectivity, and decoherence effects~\cite{d-wavesystems.2021a,d-wave.2023a}.

Prior work shown that the training of SVMs can be formulated as a QUBO problem, enabling their optimization via QA~\cite{muecke.etal.2019a,willsch.etal.2020a,date.etal.2020a}.
This formulation introduces a tradeoff between the number of available qubits and the precision with which SVM parameters can be represented.
Specifically, increasing parameter precision requires more qubits, potentially exceeding current hardware capacities.

In this work, we investigate the relationship between the number of qubits---determining the discretization level of SVM parameters---and the resulting predictive performance on various datasets with varying number of features and data points. 

Furthermore, we compare the \QUBO-based SVM training with LIBSVM, a widely-used classical solver that employs full floating-point precision~\cite{chang.lin.2011a}.
While higher precision is theoretically expected to yield better model performance, our experiments show that even low-precision QUBO-based training can produce models with competitive, and in some cases superior, predictive accuracy.

\section{Background}

We briefly summarize the theoretical tools used throughout this article, namely SVMs and \QUBO.
As notational conventions, we use $\BB=\set{0,1}$, $\BS=\set{+1,-1}$, and $\range{n}=\lbrace 1,\dots,n\rbrace$.
The vector $\bm{1}_n$ denotes $[1,1,\dots,1]^\T\in\BR^n$, $\bm{0}_n$ analogously denotes $[0,0,\dots,0]^\T\in\BR^n$, and $\bm{I}_n$ is the $n\times n$ identity matrix.
The element-wise product of two vectors or matrices denoted by $\odot$, and the Kronecker product by $\otimes$.

\subsection{Support Vector Machines}

Given a data matrix $\bX\in\BR^{N\times d}$ and corresponding binary labels $\by\in\set{+1,-1}$, we want to find a hyperplane in $d$-dimensional space defined by its normal vector $\bw\in\BR^{d}$ and a bias term $b\in\BR$, such that \begin{enumerate*}[label=(\roman*)]\item all points on the same side of have the same label, and \item the point-free margin around the hyperplane is maximized\end{enumerate*}.
As perfect separation is rarely possible in practice, we penalize points on the wrong side of the hyperplane.
This leads to the following \emph{primal} optimization problem:
\begin{definition}[Primal SVM~\cite{hastie.etal.2009a}]\label{def:primalsvm}
	Given a labeled data set $\cD=\lbrace(\bx^i,y_i)\rbrace_{i\in\range{N}}\subset\BR^d\times\BS$ and a feature map $\varphi:\BR^d\rightarrow\BR^f$, the \emph{primal SVM} is the optimization problem \begin{align}\label{eq:primalsvm}
		\underset{\bm w,b}{\min} ~&\dfrac{1}{2}\norm{\bm w}_2^2+C\bm{1}_N^\T\bm\xi\\
		\text{s.t.} ~&(\bm w^\T\varphi(\bx^i)-b)y_i\geq 1-\xi_i ~\forall\,i\in\range{N},\nonumber
	\end{align}
	where $\bm\xi=(\xi_1,\dots,\xi_N)^\T$, $\xi_i=\max(0,1-y_i(\bm{w}^\T\varphi(\bx^i)-b))~\forall\,i\in\range{N}$, and $C>0$ is a hyperparameter controlling the impact of misclassification.
\end{definition}

Using Lagrangian optimization, we obtain an equivalent dual problem:

\begin{definition}[Dual SVM~\cite{hastie.etal.2009a}]
	Let $\cD$ as in \cref{def:primalsvm}, with $\by\in\BR^N$ the vector containing all $y_i$ in $\cD$, $\bX$ the $N\times d$ data matrix, and $\bm K=\varphi(\bX)^\T\varphi(\bX)$ the \emph{kernel matrix} w.r.t. some feature map $\varphi$.
	The dual form of the SVM training problem given in \cref{def:primalsvm} is \begin{align*}
		\underset{\bm\alpha}{\text{Maximize}} ~&\bm{1}_{N}^\T\bm\alpha-\dfrac{1}{2}\bm\alpha^\T(\by\by^\T\odot\bm K)\bm\alpha\\
		\text{s.t.} ~&0\leq\alpha_i\leq C ~\forall\,i\in\range{N},\\
		&\bm\alpha^\T\by=0.
	\end{align*}
\end{definition}

Once the Lagrangian dual problem of the SVM has been solved, yielding the optimal dual variables $\bm{\alpha}^* \in \mathbb{R}^N$, the weight vector $\bw$ of the primal solution can be recovered directly.
Specifically, the optimal weights are given by $\bw = \sum_{i=1}^N \alpha_i^* y_i \bx^i$, where $\alpha_i^*$ is the optimal dual variable corresponding to training example $\bx^i$ with label $y_i \in \BS$.
This expression follows from the KKT conditions, which ensure that $\bw$ lies in the span of the training data.

The bias term $b$ can be recovered using any support vector $\bx^j$ for which $0 < \alpha_j^* < C$.
For such a support vector, the complementary slackness condition implies $b = \bw^\T\bx^j-y_j$.
In practice, the bias is often computed by averaging over all support vectors that satisfy $0 < \alpha_i^* < C$ to improve numerical stability.

\subsection{QUBO}

The \QUBO problem, (short for \emph{Quadratic Unconstrained Binary Optimization}) is a versatile class of optimization problems defined as follows:

\begin{definition}[\QUBO]
	Let $\bQ\in\BR^{n\times n}$ a symmetrical matrix for a fixed $n\in\BN$.
	The \emph{energy function} $E_{\bQ}:\BB^n\rightarrow\BR$ is defined as \begin{equation*}
		E_{\bQ}(\bz)=\bz^\T\bQ\bz=\sum_{i=1}^n\sum_{j=1}^nQ_{ij}z_iz_j,
	\end{equation*}%
	where $\BB=\set{0,1}$ denotes the set of binary digits.
	\QUBO is the problem of finding a binary vector $\bz^*$ that minimizes $E_{\bQ}$, i.e., $\forall\bz\in\BB^n:~E_{\bQ}(\bz^*)\leq E_{\bQ}(\bz)$.
\end{definition}

It is closely related and computationally equivalent to the \emph{Ising model}, which uses binary variables in $\BS$ instead of $\BB$.
Problems of either form can be converted into each other by applying the identity $\sigma = 1-2z$ for $\sigma\in\BS$ and $z\in\BB$.
Numerous optimization problems have been re-formulated as \QUBO~\cite{lucas.2014a,date.etal.2020a,muecke.etal.2023a,muecke.gerlach.2023a,muecke.2024a}.
Its great value lies in its compatibility with quantum annealing (QA), a type of adiabatic quantum computing (AQC) which can find the ground state of Ising models by exploiting quantum effects~\cite{kadowaki.nishimori.1998a}, and therefore find minimizing solutions of \QUBO problems.
Although physical quantum computers are still at an early stage in their development~\cite{preskill.2018a}, QA holds the promise to solve \QUBO and the Ising model very efficiently, even for large numbers of variables.
Nevertheless, quantum annealers can be used today to solve \QUBO problems with a limited number of variables and number of non-zero values in $\bQ$~\cite{d-wavesystems.2021a}, and strategies to use this early quantum hardware effectively are the subject of current research~\cite{muecke.etal.2024a}.
Moreover, various classical (i.e., non-quantum) solution or approximation algorithms have been devised~\cite{kochenberger.etal.2014a,punnen.etal.2022a}.

\section{A QUBO Formulation of SVM Training}

For simplicity, we focus on the linear SVM by fixing the feature map $\varphi$ to the identity function, leading to $\bm K=\bX^\T\bX$ being the Gram matrix.
If we take a closer look at the equation, we find that it has strong similarity with \QUBO, although with a few differences, namely \begin{enumerate*}[label=(\roman*)]\item we have a maximization instead of a minimization problem, \item the values $\bm\alpha$ that we optimize are real-valued, not binary, and \item there are additional constraints on $\alpha$\end{enumerate*}.
Firstly, we can flip the sign to convert a maximization to a minimization problem.
The second point, however, is more challenging, because \QUBO's domain is binary variables $\bz\in\BB^N$.
To resolve this, we can take a radical approach and binarize the elements of $\bm\alpha$ and simply set $\alpha_i=Cz_i$, meaning that $\alpha_i$ can take \emph{either} $0$ or $C$ instead of being a real value in the interval $[0,C]$.
By doing this, we automatically enforce the first set of constraints, namely $0\leq\alpha_i\leq C$.
The second constraint, $\bm\alpha^\T\by=0$, can be enforced by adding a penalty term \begin{equation*}
	\lambda(\bm\alpha^\T\by)^2 = \lambda\cdot\bm\alpha^\T\by\by^\T\bm\alpha,
\end{equation*}
which is 0 when the condition is fulfilled, and assumes a positive value otherwise.
We have to choose a value $\lambda>0$ that is large enough to make any solution that violates the constraint non-optimal.

Combining these ideas, we arrive at the following \QUBO formulation of the binarized SVM learning problem:
\begin{definition}[Binary \QUBO-SVM~\cite{muecke.etal.2019a}]\label{def:qubosvm}
	Let $C,\lambda>0$, and let $\bX\in\BR^{N\times d}$ and $\by\in\BS^N$.
	The Binary \QUBO-SVM can be trained by solving the \QUBO problem given by \begin{equation*}
		E_{\mathrm{SVM}}(\bz; C,\lambda,\bX,\by) = -\bm{1}_{N}^\T\bz+C\bz^\T\biggl(\frac 1 2(\by\by^\T\odot\bm K)+\lambda\by\by^\T\biggr)\bz.
	\end{equation*}
	Given a minimizer $\bz^*\in\BB^N$, the parameters $\bm w$ and $b$ used for the prediction function are then given by \begin{align*}
		\bm w &= C\cdot(\bz^*\odot\by)^\T\bX = C\sum_{i=1}^Nz_i^*y_i\bx^{i\T}, &
		b &= \bm w^\T\bx^{i\T}-y_i ~\text{for any }i\in\range{N},
	\end{align*}
	where $\bx^i$ is the $i$-th row of the data matrix, i.e., the $i$-th data point as a row vector.
\end{definition}

Conveniently, the solution vector $\bz^*$ to the \QUBO problem is at the same time an indicator vector of which data points are support vectors, i.e., which vectors contribute to the decision boundary.

\subsection{Increasing Precision}

The assumption of $\alpha_i=Cz_i$ is quite radical in that it allows no gradation between being no support vector at all ($\alpha_i=0$) or being misclassified ($\alpha_i=C$).
From another point of view, data points can either not contribute to the optimal $\bm w$ at all, or equally with weight $C$.

Several authors have introduced the idea of softening this restriction by encoding the weights $\alpha_i$ using $k>1$ bits instead of only one~\cite{willsch.etal.2020a,date.etal.2020a}.
Assume we have a vector $\bm p=(p_1,\dots,p_k)^\T\in\BR_{0+}$ with $\norm{\bm p}_1=C$.
We can represent each $\alpha_i$ as a sum $\sum_{j=1}^kp_jz_{i,j}$ with $z_{i,j}\in\BB~\forall j\in\range{k}$, such that for each combination of bits that sum is bounded between $0$ and $C$.
To do this for all $\alpha_i$, let now $\bz=(z_{1,1},\dots,z_{1,k},z_{2,1},\dots,z_{N,k})^\T\in\BB^{Nk}$ be a binary vector where the $k$ bits for each $\alpha_i$ are simply concatenated.
We can construct a matrix that maps this $k$-times larger vector to an $N$-element vector utilizing the Kronecker product: \begin{equation}
	\bm P=\bm{I}_N\otimes\bm p^\T \in\BR^{N\times Nk}.
\end{equation}
This way, we obtain $\bm\alpha=\bm P\bz\in[0,C]^N$ for any $\bm p\in\BR_{0+}^k$ with $\norm{\bm p}_1=C$.
Replacing $\bm\alpha$ by $\bm P\bm x$ in \cref{def:qubosvm} yields the following definition.
\begin{definition}[$k$-bit \QUBO-SVM]\label{def:kqubosvm}
	Let $C,\lambda>0$, $\bX\in\BR^{N\times d}$ and $\by\in\BS^N$ as before.
	Further, let $\bm p\in\BR^k$ with $\norm{\bm p}_1=C$.
	The $k$-bit \QUBO-SVM can be trained by solving the \QUBO problem given by \begin{equation}\label{eq:kqubosvm}
		E_{\mathrm{SVM},\bm{k}}(\bz; C,\lambda,\bX,\by,\bm p) = -\bm{1}_{N}^\T\bm P\bz+\bz^\T\bm P^\T\biggl(\frac 1 2(\by\by^\T\odot\bm K)+\lambda\by\by^\T\biggr)\bm P\bz,
	\end{equation}
	where $\bm P=\bm{I}_N\otimes\bm p$.
	Given a minimizer $\bz^*\in\BB^{Nk}$, the parameters $\bm w$ and $b$ used for the prediction function are then given by \begin{align*}
		\bm w &= ((\bm P\bz^*)\odot\by)^\T\bX, &
		b &= \bm w^\T\bx^{i\T}-y_i ~\text{for any }i\in\range{N},
	\end{align*}
	where $\bx^i$ is the $i$-th row of the data matrix, i.e., the $i$-th data point as a row vector.
\end{definition}

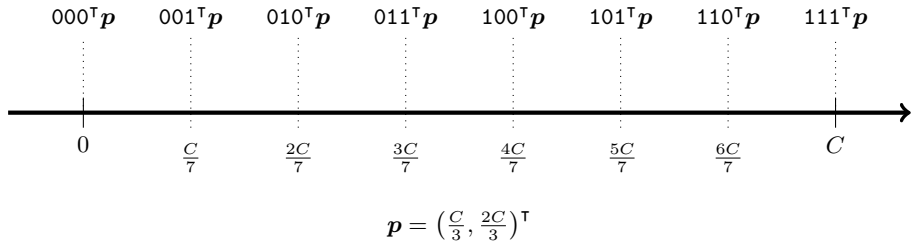
\begin{figure}[t]
	\centering
	\begin{tikzpicture}
		\draw[->,ultra thick] (-1,0) -- (11,0);
		\draw (0, 0.2) -- (0, -0.2) node[below] {$0$}
		(10, 0.2) -- (10, -0.2) node[below] {$C$};
		\draw[dotted]
		(0, 0) -- (0, 1) node[above] {$\verb|000|^\T\bm p$}
		(1.4286, -0.2) -- (1.4286, 1) node[above] {$\verb|001|^\T\bm p$}
		(2.8571, -0.2) -- (2.8571, 1) node[above] {$\verb|010|^\T\bm p$}
		(4.2857, -0.2) -- (4.2857, 1) node[above] {$\verb|011|^\T\bm p$}
		(5.7143, -0.2) -- (5.7143, 1) node[above] {$\verb|100|^\T\bm p$}
		(7.1429, -0.2) -- (7.1429, 1) node[above] {$\verb|101|^\T\bm p$}
		(8.5714, -0.2) -- (8.5714, 1) node[above] {$\verb|110|^\T\bm p$}
		(10, 0) -- (10, 1) node[above] {$\verb|111|^\T\bm p$};
		\node at (1.4286,-0.6) {$\frac{C}{7}$};
		\node at (2.8571,-0.6) {$\frac{2C}{7}$};
		\node at (4.2857,-0.6) {$\frac{3C}{7}$};
		\node at (5.7143,-0.6) {$\frac{4C}{7}$};
		\node at (7.1629,-0.6) {$\frac{5C}{7}$};
		\node at (8.5714,-0.6) {$\frac{6C}{7}$};
		\node[align=center] at (5,-1.5) {$\bm p=\left(\frac{C}{3},\frac{2C}{3}\right)^\T$};
	\end{tikzpicture}
	\caption{Example of a conversion from a binary vector of length $k=3$ to a scalar value between $0$ and $C$ through multiplication with a vector $\bm p$; by choosing $p_j=C\cdot 2^{j-1}/(2^k-1)$, the interval $[0,C]$ is sampled evenly.}
	\label{fig:precision}
\end{figure}

The question remains how to effectively choose $\bm p$.
As the encoding of the $\alpha_i$ values already resembles the base-2 number system, it is natural to choose powers of 2 to subdivide the space $[0,C]$ evenly, in a way that $\bm{0}_{k}$ represents $0$, and $\bm{1}_{k}$ represents $C$.
To achieve this, let $p_j=C\cdot 2^{j-1}/(2^k-1)$ for all $j\in\range{k}$.
It is easy to see that \begin{equation*}
	\sum_{j=1}^k\frac{C\cdot 2^{j-1}}{2^k-1}=\frac{C}{2^k-1}\sum_{j=0}^{k-1}2^j=C,
\end{equation*}
and the interval $[0,C]$ is subdivided evenly, as visualized in \cref{fig:precision}.
Another advantage of choosing this $\bm p$ is that the special case $k=1$ yields the original Binary \QUBO-SVM given in \cref{def:qubosvm}.
Therefore, we can focus on \cref{def:kqubosvm} in the following evaluation.

\subsection{Experimental Evaluation}
\label{sec:qasvm:experiments}

It remains to be investigated how the discretization of $\bm\alpha$ affects the quality of the trained classifier.
To test this, we train the $k$-bit \QUBO-SVM on a range of data sets for various values of $C$ and $k$ and record their test accuracies.
For comparison, we also train a classical linear SVM using LIBSVM~\cite{chang.lin.2011a}, which is contained as part of the scikit-learn Python package \cite{pedregosa.etal.2011a}.

\begin{table}[t]
	\caption{Data sets used for numerical experiments. For each data set, the number of features $d$ and data points $N$ is given.}
	\label{tab:svm:datasets}
	\centering
	\begin{tabular}{llcc}
		\hline\noalign{\smallskip}
		Name & References & $d$ & $N$ \\
		\noalign{\smallskip}\hline\noalign{\smallskip}
		\texttt{iris} (mod.) &\cite{fisher.1936a} &4 &150 \\
		\texttt{sonar} &\cite{gorman.sejnowski.1988a} &60 &208 \\
		\texttt{mnist} (mod.) &\cite{lecun.cortes.2010a} &196 &200 \\
		\noalign{\smallskip}\hline
	\end{tabular}
\end{table}

The data sets we use are listed in \cref{tab:svm:datasets}, all of which are widely used as classification benchmarks throughout literature.
As the SVM (in its original form) can only separate two classes, we modify both \texttt{iris} and \texttt{mnist}:
For \texttt{iris}, we only use the classes \textit{versicolor} and \textit{virginica}, which leaves $N=100$ data points.
For \texttt{mnist}, we only use 100 samples each of the digits 4 and 7 as our two classes.
To reduce the number of features, we perform max-pooling over every $2\times 2$ pixel block, reducing the image size to $14\times 14$ pixels.
We normalize the values to the interval $[0,1]$ and then linearize the images to vectors of length $196$, arriving at a data set with $N=200$ and $d=196$.

To get a wider overview over the model's performance, we test all combinations of values $C\in\lbrace 2^{-6},2^{-5},\dots,2^{4}\rbrace$ and $k\in\lbrace 1,2,3\rbrace$.
For choosing $\lambda$ we use an iterative approach where we start with $\lambda=1$, solve the \QUBO problem and check if the constraint $\bz^{*\T}\bm P^\T\by=0$ is violated.
If it is, we double $\lambda$ and try again until we find a valid solution.

As a solver, we use the MST2 multistart tabu search algorithm \cite{palubeckis.2004a} contained in the dwave-tabu Python package\footnote{\url{https://docs.ocean.dwavesys.com/projects/tabu/en/latest/}}, which performs one million restarts per run.
To further increase the solution quality, we perform 20 runs for each \QUBO instance.
In addition, we tried to solve the \QUBO problems on a D-Wave quantum annealer.
However, we found that \cref{eq:kqubosvm} produces dense weight matrices for which the system was not able to find an embedding onto its qubit topology.

\begin{figure}[t]
	\centering
	\includegraphics[width=\textwidth]{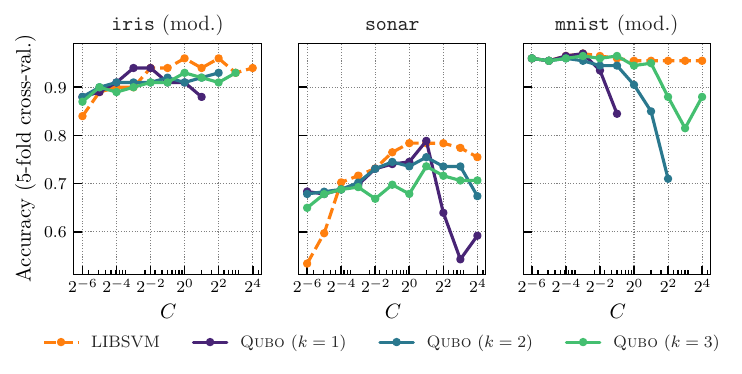}
	\caption{5-fold cross-validated prediction accuracies of the $k$-bit \QUBO-SVM and LIBSVM on the three data sets listed in \cref{tab:svm:datasets}, using various values for $C$ and $k$.}
	\label{fig:svm:testacc}
\end{figure}

We performed a 5-fold cross validation and report the mean prediction accuracy in \cref{fig:svm:testacc}, making sure to use the same splits across all models and hyperparameter settings for comparability.
Remarkably, despite the strong simplification of using even just two discrete values for all $\alpha_i$, the accuracy scores are comparable to the LIBSVM results that use floating-point values.
For small $C$, the \QUBO-SVM even surpasses LIBSVM on \texttt{iris} and \texttt{sonar}.
Surprisingly, increasing the precision by choosing a larger $k$ does not generally lead to higher accuracy; only on \texttt{mnist} a higher $k$-value leads to better solutions for larger $C$.
This implies that using a higher number of support vectors that contribute equally to the parameter vector $\bm w$ is more useful than allowing for more fine-granular weighting of fewer support vectors, which is an interesting insight, whose generality needs to be investigated more thoroughly in future work.
However, we observe that from a certain $C$ value we cannot find suitable solutions on both \texttt{iris} and \texttt{mnist}, as the minimizing vector is $\bm{0}_{Nk}$, and $\bm\alpha=\bm{0}_{N}$ accordingly.
Without any support vectors we cannot correctly compute $\bm w$, and the resulting predictions are not usable -- this is reflected by missing values in the figure.
We observe that this problem is mitigated by higher values for $k$, as we obtain usable results with higher $C$ on all data sets if we choose $k=2$ or $k=3$, at the cost of increasing the \QUBO size.

\subsection{Concluding Remarks}

SVM training can be performed using a \QUBO embedding, which can be solved on quantum annealers to obtain the set of support vectors.
The weightings $\bm\alpha$ from the dual SVM can be approximated to arbitrary precision by using $k$ bits per weight, which leads to a \QUBO formulation of size $Nk$, where $N$ is the number of data points.
We have seen that, while a higher $k$ does not necessarily increase the classification accuracy, it allows for higher values of $C$ by avoiding all-zero solutions to the \QUBO problem.
In general, however, the accuracy for lower $C$ is competitive with LIBSVM, even for $k=1$, which is a surprising observation given that LIBSVM uses full floating-point precising.
This implies that the correct choice of support vectors may be more crucial for a good performance that their exact weighting.

The size of the \QUBO problem is, at the current point of time, a limiting factor for the \QUBO-SVM, as the number of variables is $Nk$.
This limits its use to data sets with only up to a few hundred data points.
However, with further improvements in QA, it may be faster to train SVMs on large data sets using noise-free quantum annealers with a large number of qubits in the future, as classical SVM training algorithms such as SMO also scale polynomially in the number of data points~\cite{platt.1998a}.
Moreover, AQC on perfect hardware would be able to find the globally optimal set of support vectors, whereas many classical algorithms use local search methods.
Therefore AQC has the potential to contribute to more accurate classification models, once the hardware has reached a point where it can be used efficiently for large-scale problems.

\begin{credits}
\subsubsection{\ackname} This research has been funded by the Federal Ministry of Education and Research of Germany and the state of North Rhine-Westphalia as part of the Lamarr Institute for Machine Learning and Artificial Intelligence.

\subsubsection{\discintname}
The authors have no competing interests to declare that are
relevant to the content of this article.
\end{credits}

\end{document}